\documentclass[letterpaper, 10 pt, conference]{ieeeconf}  %
\IEEEoverridecommandlockouts                              %
\usepackage{graphics} 
\usepackage{graphicx}
\usepackage{amsmath} 
\usepackage{xcolor}
\usepackage{multirow}
\usepackage{bm}
\usepackage{makecell}
\usepackage{url}
\usepackage{float}
\usepackage{subcaption}
\usepackage{booktabs,makecell,multirow}
\usepackage{amsfonts}
\usepackage{mathrsfs}
\usepackage{cite}
\usepackage{color}
\usepackage{amsmath,amsfonts,amssymb}
\usepackage{amsfonts}
\usepackage{tikz}
\usetikzlibrary{positioning,fit,backgrounds,arrows.meta,calc}
\usepackage[some]{background}
\usepackage{amsmath}
\usepackage[hidelinks]{hyperref}
\newif\ifrev \revtrue

\title{\LARGE \bf
Leveraging Human-In-The-Loop Demonstrations in Reinforcement Learning for Digital Twin-Driven Robot Flexibility
}

\author{Yuzhu Sun, Mien Van, Nguyen Minh Nhat, Stephen McIlvanna and Seán McLoone\thanks{This work was supported by the European Commission through the Horizon Europe project HARMONICA under grant agreement No 101294331. Views and opinions expressed are those of the authors only and do not necessarily reflect those of the European Union or the European Commission.}\thanks{Yuzhu Sun, Mien Van (corresponding author),  Nguyen Minh Nhat, Stephen McIlvanna and Seán McLoone are with the School of Electronics, Electrical Engineering and Computer Science, Queen's University Belfast, Northern Ireland, UK. E-mail: {\tt\small ysun32@qub.ac.uk; m.van@qub.ac.uk; nnhat01@qub.ac.uk; smcilvanna0@qub.ac.uk; s.mcloone@qub.ac.uk}} %
}

\begin{document}

\maketitle
\thispagestyle{empty}
\pagestyle{empty}

\begin{abstract}

Growing automation makes collaborative robots work in more variable environments, increasing the need for adaptation. We propose a human-in-the-loop online training framework combining a digital twin (DT), reinforcement learning (RL), and human demonstrations. Unlike DTs used mainly to generate synthetic data before task execution, our DT is synchronized with the physical system in real time through camera feeds, allowing the virtual robot to update its observations and policy from real-world feedback. A dual actor framework integrates imitation learning (IL) without adding a direct imitation loss to the RL actor, so demonstrations can guide adaptation instead of manual reprogramming. The proposed framework is demonstrated on the Ufactory Xarm5 collaborative robot, where the robot's end-effector aims to reach the target position while avoiding obstacles. The experiments show that the framework can resume training after a change in the physical workspace and that, with a fixed set of non-optimal demonstrations, the dual actor framework achieves a much higher final success rate than two methods that add an imitation loss to the actor. The same pattern holds with real human demonstrations collected in virtual reality (VR): with demonstrations that never reach the goal, the dual actor framework reached 83--100\% mean deterministic evaluation success, against 0--17\% for the two imitation-loss methods.


\end{abstract}

\section{INTRODUCTION}\label{sec1}
With growing automation in Industry 5.0, sensor-equipped collaborative robots increasingly work alongside people rather than behind safety cages. More complex and unpredictable workspaces therefore require greater adaptability and flexibility \cite{Malik2020ManMA}.
Artificial intelligence (AI), particularly reinforcement learning (RL), can improve robot autonomy by learning through trial and error from environmental feedback \cite{RN313}.
Training RL agents directly on physical robots is costly, potentially hazardous, and often impractical.
A digital twin (DT) represents the physical workspace in simulation, allowing an RL agent to train and receive feedback without repeated physical trials.
Digital twins have often been used to generate synthetic data or provide simulated environments for training perception and control models before deployment \cite{RN391,RN402,RN372,RN396}. Sim-to-real transfer methods seek to bridge the remaining differences between these virtual environments and the physical robot \cite{8202133,RN401}. This separation between training and operation becomes a problem when the workspace changes: the twin may no longer reflect the physical scene, and the model or policy may need to be updated \cite{RN404}. Such updates can interrupt task execution.

We address this gap with a PyBullet digital twin of the Ufactory Xarm5 \cite{coumans2021}, linked to the physical robot through the Robot Operating System (ROS). Camera feeds keep the twin synchronized with the workspace in real time. If the twin detects that the current policy fails an obstacle-avoidance task after a workspace change, the RL agent resumes training in simulation.

\begin{figure*}[t]
	\centering
	\includegraphics[width=0.8\textwidth]{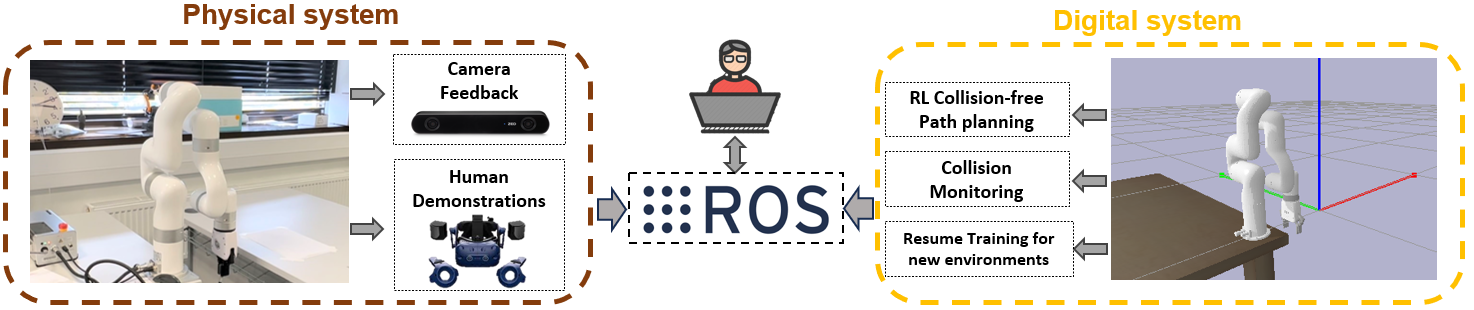}
	\caption{The structure of the proposed system.}
	\label{system}
\end{figure*}

During this resumed training, an operator can demonstrate the desired behaviour to guide the robot's adaptation. This lets the operator contribute task knowledge without manually reprogramming the robot or needing specialist AI expertise.
Previous work has used demonstrations in offline reinforcement learning \cite{RN434, RN452} and in methods combining imitation learning (IL) with RL \cite{RN431, RN441, RN442}. Many studies use datasets generated by pre-trained experts, which help evaluate algorithms but differ from demonstrations collected from operators during deployment. For value learning, demonstrated transitions also need reward labels \cite{RN452}; raw human motion data do not provide these automatically. This makes it harder to use operator demonstrations during the resumed training described above.
To address this, we capture the operator's movements in virtual reality (VR) and transfer them to the DT, where they guide the virtual robot. The DT evaluates each demonstrated step using the task reward and labels the resulting transitions automatically. The same environment therefore supports both resumed RL training and the preparation of human demonstrations.

The DT provides reward labels for human demonstrations, but it cannot make the demonstrated behaviour optimal. In our experiments, methods that combine IL and RL through an imitation loss on the RL actor \cite{RN439, RN442, RN440, RN441} performed poorly with noisy, non-optimal demonstrations. The imitation loss encourages the actor to repeat those actions, potentially limiting its exploration of better solutions.
To address this limitation, we use two actors with distinct roles: an RL actor learns from interactions with the DT, while an IL actor learns from human demonstrations. Both paths contribute to the critic target, which favours the one with higher estimated value. With non-optimal demonstrations, the proposed framework reached over 80\% mean final training success, while two imitation-loss baselines rarely completed the task. Fig.~\ref{system} shows the framework. Our main contributions are:

\begin{enumerate}
\item A hardware-in-the-loop DT that stays synchronized with the physical workspace and resumes RL training when a change causes the current policy to fail. Unlike DTs used mainly for synthetic data or training before deployment \cite{RN402, RN372, RN396, RN391}, it updates the learned policy during operation until the task can again be completed.
\item A dual actor framework in which demonstrations train a separate behaviour-cloning actor and guide the RL actor only through the critic target. Without a direct imitation loss, the RL actor can explore beyond poor demonstrations.
\item An evaluation with real human VR demonstrations that the DT labels with task rewards automatically. We report collection effort and workload; with complete demonstrations, the dual actor framework converged in all six seeds, compared with at most one for the imitation-loss baseline methods. We also test retraining in three additional modified scenes.
\end{enumerate}

The remainder of this paper is organized as follows. Section~\ref{sec2} reviews related work. Section~\ref{sec3} describes the DT, and Section~\ref{sec4} details the dual actor framework. Section~\ref{sec5} presents the experiments and discussion; Section~\ref{sec6} concludes the paper.

\section{Related Work}\label{sec2}
\subsection{Reinforcement Learning for Robotic Manipulation}
RL has been used to learn robot behaviour from reward feedback \cite{RN412}, with applications in path and motion planning \cite{RN334,RN322,RN348,RN389}, task allocation \cite{RN321}, human-robot collaboration \cite{RN312,RN386}, and learning from demonstrations \cite{RN323}. For collaborative robots, adapting a learned policy as people and workspaces change is particularly important. In \cite{RN387}, RL balances human avoidance with task efficiency using an avatar animated from camera-based skeleton data. In \cite{RN313}, goal-conditioned RL combines collision avoidance and contact-rich tasks while the co-worker's hand follows random trajectories. Both \cite{RN387} and \cite{RN313}, however, train the policy before deployment and do not include a mechanism that detects a change in the environment during operation and resumes training in response. In contrast, this paper investigates hardware-in-the-loop retraining combined with DT technology, in which training is resumed during task execution whenever the DT detects that the environment has changed. This shifts the human effort from re-programming the robot to simply providing a few demonstrations.

\subsection{Digital Twin}
DTs provide virtual representations of physical systems \cite{RN413}. Since Grieves first formulated the DT concept in the context of product lifecycle management \cite{grieves2014digital}, the term has broadened from a static virtual replica into a decision-support layer that is expected to monitor, predict and optimize its physical counterpart \cite{RN414}. A twin that is calibrated once at design time and not updated during operation can, however, become inaccurate when the environment changes, which limits its value precisely in the situations where an autonomous robot needs it most. Their connection to physical systems makes simulation useful for developing and testing robotics methods without repeated physical trials. Applications include grasping \cite{RN395,RN370}, navigation \cite{8247712}, obstacle avoidance \cite{RN417}, and task allocation \cite{RN381,RN416}.

For machine learning, DTs often provide synthetic data or simulated training environments before a robot begins its task. Some of these twins are synchronized with the physical system in real time \cite{RN402, RN372}, others only in part \cite{RN391, RN396}, and the grasp-detection model in \cite{RN402} is updated by continual learning when new objects are introduced; these updates, however, are offline training sessions on collected data rather than training resumed during task execution. After training, workspace changes can leave the DT out of sync with the physical robot and require remodelling or reprogramming. In \cite{RN410}, video feeds update a bridge-inspection robot’s DT so its path can be replanned online. The DT in \cite{RN410}, however, drives an online \emph{planner} rather than a learning agent, and the accuracy of the twin itself can drift over time after deployment. Building on this idea, this paper couples a continuously synchronized DT directly to the RL training loop, so that policy learning itself, and not only planning, reacts to real-time changes in the environment.

\subsection{Integrating demonstrations into RL}
Demonstrations can guide the RL updates made possible by a synchronized DT. Initializing or guiding RL with supervised learning from demonstrations is an established idea \cite{RN487}. Prior studies have used them to improve exploration and sample efficiency \cite{RN451, RN487, RN485, RN486, RN439, RN484, RN477, RN435}, but differ in how the data enter the learner.

Methods include (i) retaining demonstrations in a prioritized replay buffer \cite{RN485}, (ii) pre-training with IL or offline RL before fine-tuning with RL \cite{RN434, RN487, RN431}, and (iii) adding an IL loss to the RL actor objective \cite{RN439, RN442, RN440, RN441, RN487}. A fourth approach selects between human teleoperation and autonomous controllers for each episode based on predicted success, with one controller learned online from demonstrations and RL \cite{RN477}. In \cite{RN485}, a simulated arm tracks a guided physical arm, and the demonstrations remain in a replay buffer alongside the agent's experience; the effect of non-optimal demonstrations was not evaluated. In \cite{RN487}, behaviour-cloning pre-training is followed by policy-gradient fine-tuning with a demonstration term whose weight decays. These training arrangements differ from ours, in which RL training resumes when the synchronized DT detects a change during operation.

Another line of work learns from demonstrations without manually assigned task rewards. Adversarial imitation learning \cite{ho2016generative} and inverse soft-Q learning \cite{garg2021iqlearn} learn from raw trajectories, while SQIL \cite{reddy2020sqil} assigns a fixed reward to demonstrated transitions. These methods use demonstrations as examples of the desired behaviour. AWAC \cite{RN434}, by contrast, uses reward-labelled offline data, including suboptimal transitions, and improves the policy through online interaction. In our framework, raw VR motion is replayed in the synchronized DT, which assigns the task reward to every transition. This allows an actor-critic learner to use human demonstrations without manual reward annotation. Demonstrations recorded in simulation are labelled automatically in \cite{RN485,RN439} as well; here, the same synchronized DT also supports training resumed during operation.

Methods such as \cite{RN439, RN442, RN440} add an auxiliary imitation loss directly to the RL actor. The Q-Filter in \cite{RN439} addresses suboptimal demonstrations by disabling imitation of actions the critic values below the actor's own actions. We therefore compare our dual actor framework with imitation-loss baselines with and without this filter. Table~\ref{tab:positioning} summarizes their differences in online adaptation, DT synchronization, and treatment of non-optimal demonstrations.

\begin{table}[t]
	\caption{Positioning of the proposed framework relative to representative approaches. Online adapt.: training resumed during operation when a change in the environment is detected; partial: holds only in part; --: not provided or not evaluated; n/a: not applicable.}
	\label{tab:positioning}
	\centering
	\footnotesize\ifrev\color{black}\fi
	\setlength{\tabcolsep}{2pt}
	\begin{tabular}{lccc}
		\toprule
		Approach & \makecell{Online\\adapt.} & \makecell{Real-time\\DT sync} & \makecell{Robust to\\non-opt.\ demos} \\
		\midrule
		Synthetic-data DT \cite{RN391,RN396} & -- & partial & n/a \\
		Synchronized DT \cite{RN372} & -- & \checkmark & n/a \\
		DT + continual learning \cite{RN402} & partial & \checkmark & n/a \\
		Prioritized demo buffer \cite{RN485} & -- & partial & -- \\
		BC pre-train + demo-aug.\ PG \cite{RN487} & -- & -- & -- \\
		Offline-to-online RL \cite{RN434} & -- & -- & \checkmark \\
		IL loss + Q-Filter \cite{RN439} & -- & -- & partial \\
		Human or agent selection \cite{RN477} & partial & -- & partial \\
		Unlabeled IL \cite{ho2016generative,reddy2020sqil} & -- & -- & -- \\
		\textbf{Proposed} & \checkmark & \checkmark & \checkmark \\
		\bottomrule
	\end{tabular}
\end{table}

\section{Digital Twin Online Training Framework}\label{sec3}
In this section, we provide details on the proposed hardware-in-the-loop training framework, including hardware setup, sensor perception, simulator development, bidirectional data transmission, and the integrated DT closed-loop system.

\subsection{Digital Twin Setup}
\subsubsection{Hardware} 
The physical setup includes the Ufactory Xarm5 robot, a ZED 2i depth camera, physical obstacles and goal objects, and an HTC VIVE system. The camera captures images and depth data for locating objects; and the VR controller lets an operator guide the virtual robot's end-effector when collecting demonstrations. The obstacles and goals provide physical tests of task performance. Because detections are initially expressed in camera coordinates, we map them into PyBullet world coordinates so the twin reflects the physical scene.

\begin{figure}[h]
	\centering
	\includegraphics[width=\linewidth]{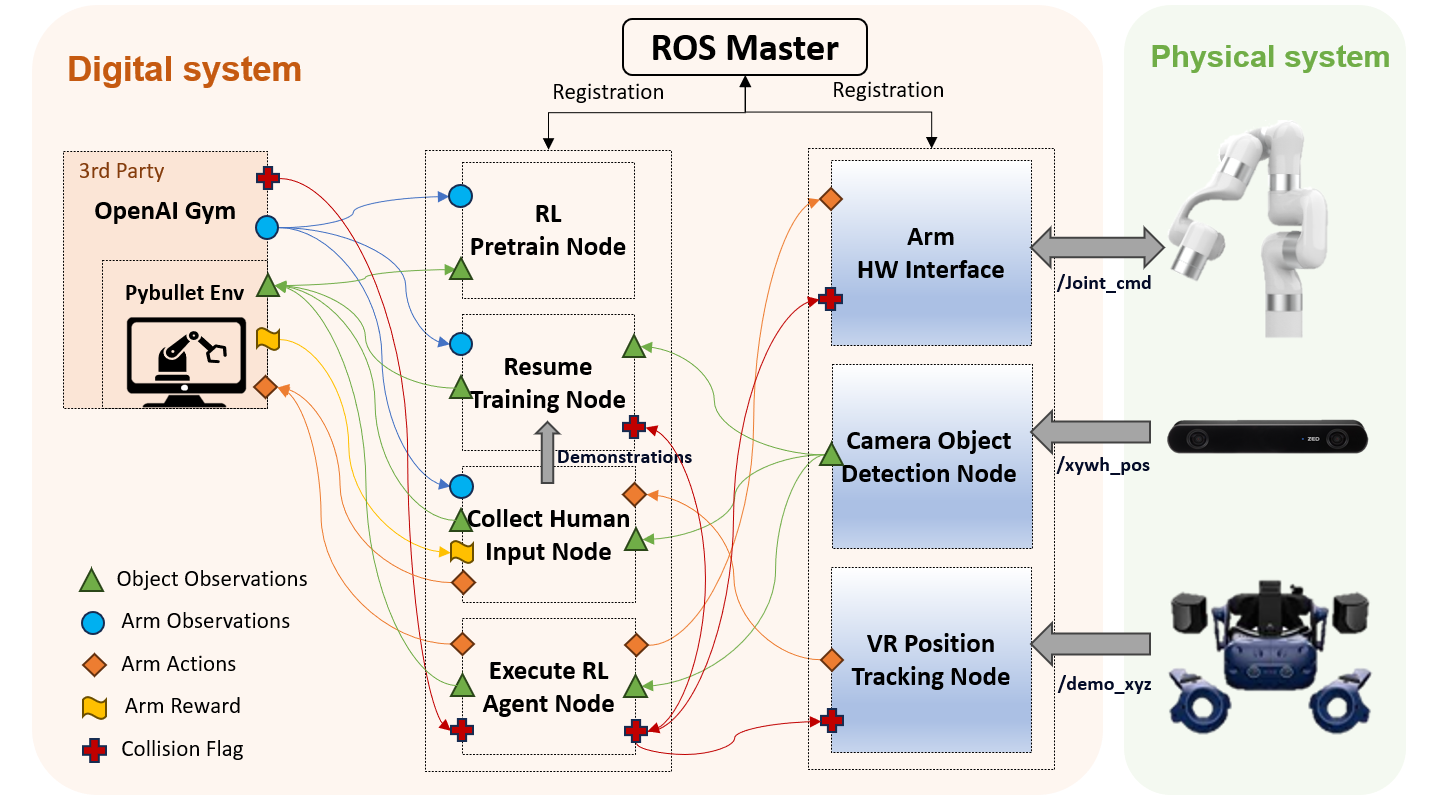}
	\caption{The Overview of digital twin software setup.}
	\label{ros}
\end{figure}

\subsubsection{Software} 
The digital model is built in PyBullet, which uses the Bullet physics engine. We implement the task as an OpenAI Gym environment, defining states, actions, and rewards through a modular interface so compatible RL algorithms can be tested on the same task. ROS exchanges data between the physical and virtual components through nodes and topics. Fig.~\ref{ros} summarizes these connections and their messages. The next subsection explains how camera detections are located in the simulated workspace.

\subsection{Object Detection and Localization}
\begin{figure}[t]
	\centering
	\includegraphics[width=0.45\textwidth]{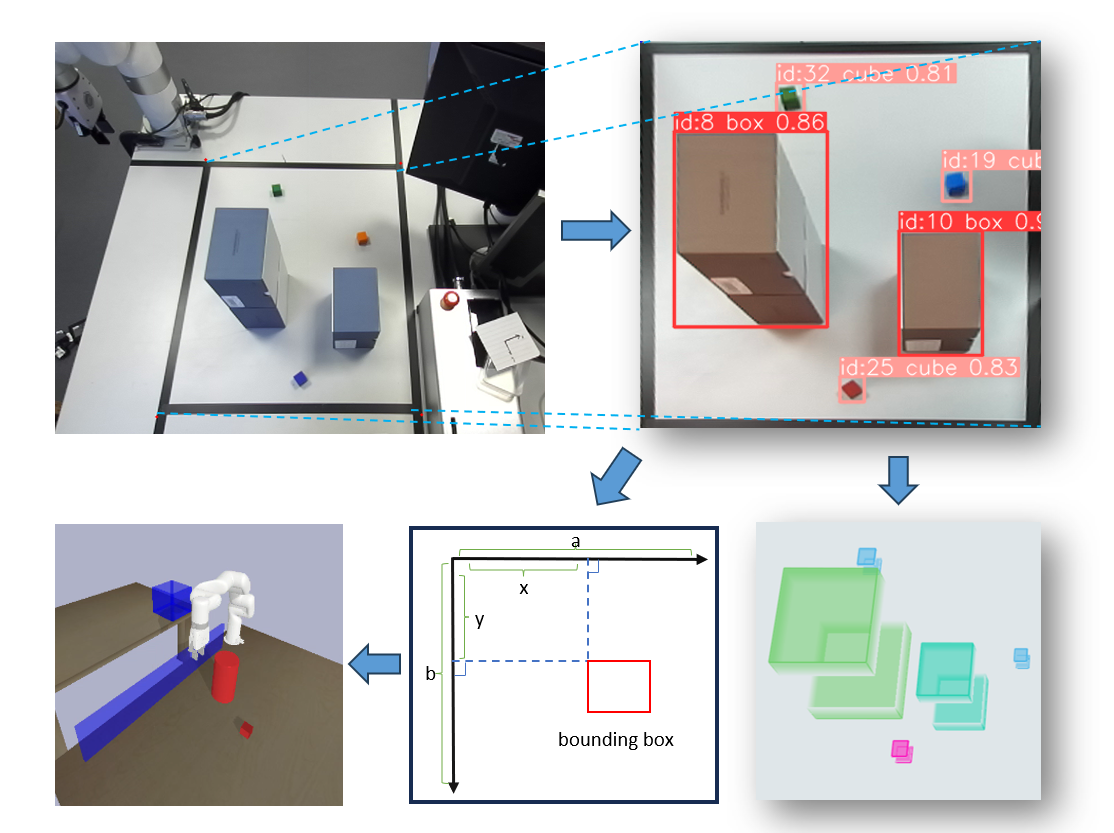}
	\caption{Object detection results.}
	\label{yolo}
\end{figure}

Camera images update the DT in two stages: we detect goals and obstacles, then map their positions into PyBullet coordinates. We use YOLOv8 \cite{Jocher_Ultralytics_YOLO_2023}, trained on 400 workspace images with a mean average precision (mAP) of $99.4\%$; the dataset is available at \cite{WinNT}. Each detection provides a class and a 2D bounding box $(x,y,w,h)$. Combined with ZED depth data, this gives the object's position and height in the physical workspace. We calibrate an OpenCV perspective transformation using the four corners of the taped workspace, producing a rectified view (Fig.~\ref{yolo}). Pixel positions in this view are scaled and offset to PyBullet world coordinates:

\begin{equation}
	\begin{cases}
		x=-r_x\left( \frac{x_p}{p_x} \right) +\varDelta x\\
		y=r_y\left( \frac{y_p}{p_y} \right) +\varDelta y\\
		z=\frac{h_z}{2}\\
	\end{cases}
\end{equation}

here, \(x_p\) and \(y_p\) are detection pixel coordinates in the transformed view; \(r_x\) and \(r_y\) are the physical lengths of the taped workspace; and \(p_x\) and \(p_y\) are the corresponding image dimensions in pixels. The measured offsets \(\Delta x\) and \(\Delta y\) align the coordinate frames. With the table at \(z=0\), the centre of an object of height \(h_z\) is at \(z=h_z/2\).

The linear form of Eq.~(1) rests on the following assumptions: (i) the camera pose is fixed with respect to the workspace, and the perspective transformation removes perspective distortion, so pixel coordinates in the transformed view are proportional to metric coordinates on the table plane; (ii) all objects rest on the table plane, so the height $h_z$ fully determines the $z$ coordinate of an object's center; and (iii) residual lens distortion after calibration is negligible over the taped workspace area. Under these conditions, the pixel-to-world mapping reduces to a per-axis scaling plus a constant offset, calibrated once from the four workspace corners. If the camera were moved or tilted, or if objects were stacked, the transformation would need to be re-estimated or replaced by a full extrinsic calibration.

\subsection{Hardware-in-the-loop Training}
\begin{figure}[h]
	\centering
	\includegraphics[width=0.4\textwidth]{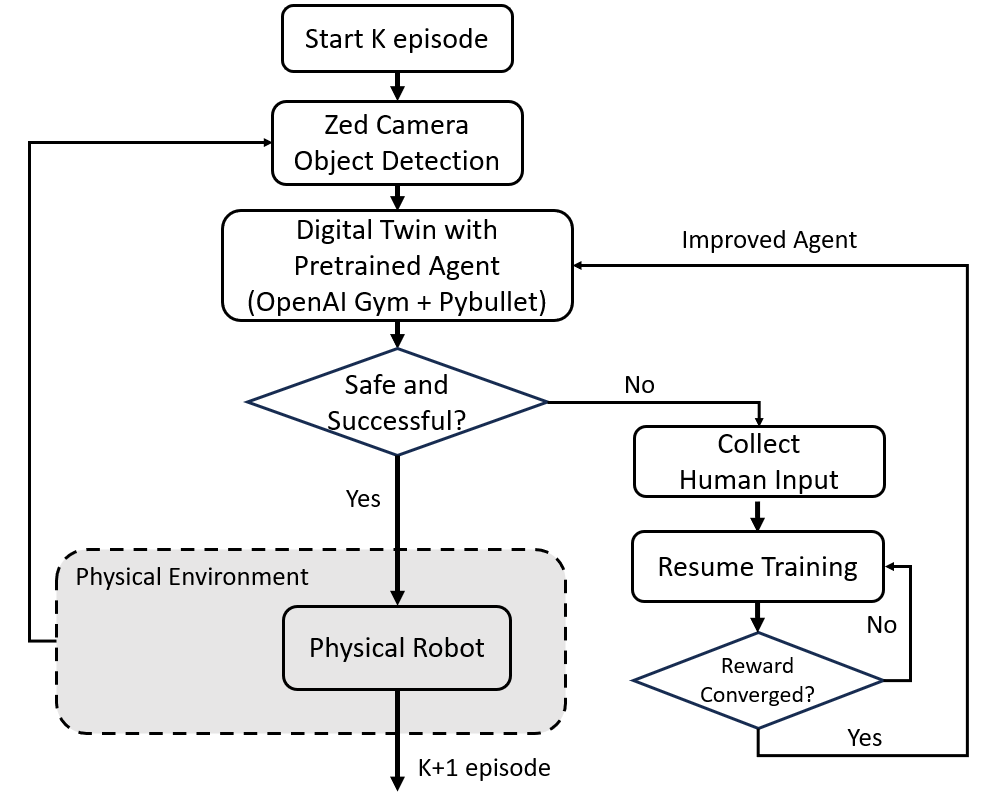}
	\caption{The flowchart of proposed framework. }
	\label{flowchart}
\end{figure}

The agent is first pre-trained in the DT before deployment. The pre-training objective is identical to the task objective used throughout the paper: the agent maximizes the reward function defined in Section~\ref{sec4} (reaching the goal while keeping a safe distance from the obstacle), with the goal and a small obstacle randomly repositioned in the workspace at every episode. Pre-training aims to produce a policy that reliably solves the initial deployment task, but this policy is not required to be optimal or to generalize to unseen environments: the only assumption the framework places on the pre-trained agent is that it achieves a stable success rate on the initial task, since any later environmental change is handled by retraining rather than by the pre-trained policy itself. Throughout the paper, we use the following terminology: \emph{pre-training} denotes training completed before deployment, \emph{retraining} denotes training resumed from the pre-trained model after the DT detects an environmental change, and the \emph{online training framework} denotes the overall closed loop shown in Fig.~\ref{flowchart}. During operation, ROS exchanges data between the physical and digital systems through nodes for camera detection, robot control, VR tracking, pre-training, retraining, and demonstration collection.

The virtual robot runs one episode ahead of the physical robot, previewing the task in about 1--2 seconds depending on the computer. If the simulation succeeds without collision, its joint commands are sent to the physical robot. If the virtual robot cannot reach the goal or hits an obstacle, camera observations update the DT, an operator provides demonstrations, and training resumes from the pre-trained model until the reward converges again. The updated policy is then tested in the DT; joint commands are sent to the physical robot only after a safe, successful simulation. Section~\ref{sec4} explains how the demonstrations guide this retraining.



\section{Integrating Human Demonstrations into Reinforcement Learning}\label{sec4}
\subsection{Preliminaries}
\subsubsection{Markov Decision Process}
Reinforcement learning relies on an agent learning how to perform a specific goal by directly interacting with its environment through trial and error. It uses the framework of Markov decision processes (MDPs) which mathematically structures the essential features of the artificial intelligence problem \cite{RN327}. The learner to be trained is called the agent and the surroundings outside of the agent are called the environment. During the interaction, the agent receives the environment's state $s_t \in S$ and selects an action $a_t \in A$ based on the transition probability $P\left( s_{t+1}|s_t,a_t \right)$ at the time step $t$, where $S$ and $A$ are the state and action spaces. As the consequence of $a_t$, the agent receives a feedback reward $r_{t+1}\in R\subset \mathbb{R} $ and finds itself a new state $s_{t+1}$ from the environment. An MDP is thereby usually represented as the following tuple: 
\begin{equation}
	\left[ S,A,P\left( s_{t+1}|s_t,a_t \right) ,R\left( s_t,s_{t+1},a_t \right) ,\gamma \right] 
\end{equation}
The overarching objective of the agent is to maximize the cumulative rewards received from environment $G_t=\sum\nolimits_{k=0}^{\infty}{\gamma ^kr_{t+k+1}}$, where $0\leqslant \gamma \leqslant 1$ is a discount factor. A higher $\gamma$ emphasis on long-term rewards and a lower $\gamma$ tends to encourage agents to pursue immediate rewards.

\subsubsection{Soft Actor Critic}
Soft Actor-Critic (SAC) is a model-free, off-policy reinforcement learning algorithm that was first proposed in \cite{RN437}. The core idea of SAC is to maximize not only the lifetime rewards but also the entropy of the policy to encourage exploration. Therefore, the objective of SAC is to maximize:

\begin{equation}
J\left( \pi \right) =\sum_{t=0}^T{\mathbb{E} _{\left( s_t.a_t \right) \sim \rho _{\pi}}\left[ r\left( s_t,a_t \right) +\alpha \mathcal{H} \left( \pi \left( \cdot |s_t \right) \right) \right]}
\label{sac1}
\end{equation}
where parameter \(\alpha\) is called the temperature parameter that determines the importance of the entropy in the optimization process. To optimize the objective function Eq. (\ref{sac1}), SAC uses neural networks to approximate three functions: a state value function \(V_{\psi}(s_t)\), a soft Q-function \(Q_{\theta}(s_t, a_t)\), and a policy function \(\pi_{\phi}(a_t | s_t)\), parameterized by \(\psi\), \(\theta\), and \(\phi\), respectively. The update rules of these three parameter vectors are to minimize the following three objective functions \(J_V\left( \psi \right)\), \(J_Q\left( \theta \right)\), and \(J_{\pi}\left( \phi \right)\):

\begin{equation}
\begin{aligned}
	J_V\left( \psi \right) =&\,\mathbb{E} _{s_t\sim \mathcal{D}}\Big[ \tfrac{1}{2}\big( V_{\psi}\left( s_t \right)\\
	&-\mathbb{E} _{a_t\sim \pi _{\phi}}\left[ Q_{\theta}\left( s_t,a_t \right) -\log \pi _{\phi}\left( a_t|s_t \right) \right] \big) ^2 \Big]\\
	J_Q\left( \theta \right) =&\,\mathbb{E} _{\left( s_t,a_t \right) \sim \mathcal{D}}\Big[ \tfrac{1}{2}\big( Q_{\theta}\left( s_t,a_t \right) -\hat{Q}\left( s_t,a_t \right) \big) ^2 \Big]\\
	J_{\pi}\left( \phi \right) =&\,\mathbb{E} _{s_t\sim \mathcal{D} ,\epsilon _t\sim \mathcal{N}}\big[ \log \pi _{\phi}\left( f_{\phi}\left( \epsilon _t;s_t \right) |s_t \right)\\
	&-Q_{\theta}\left( s_t,f_{\phi}\left( \epsilon _t;s_t \right) \right) \big]\\
\end{aligned}
	\label{sac2}
\end{equation}
\noindent where
\begin{equation}
	\hat{Q}\left( s_t,a_t \right) =r\left( s_t,a_t \right) +\gamma \mathbb{E} _{s_{t+1}\sim \rho}\left[ V_{\bar{\psi}}\left( s_{t+1} \right) \right]
	\label{sac3_1}
\end{equation}

The above three objective functions constitute the core idea of SAC. For brevity, implementation details are omitted here. More details can be found in \cite{RN437}.

\subsection{RL based Obstacle Avoidance}
\subsubsection{State}
In this paper, the objective of the robot is to reach the goal (a cube) and avoid obstacles (boxes with different sizes). Therefore, the relevant observations for the training process include the current time step $t$, the position of the cube in the $x-y$ plane $pos_{goal}$, the size of the obstacles $\left[h, w\right]$, the position $pos_{tcp}$ and velocity $vel_{tcp}$ of the robot Tool Center Point (TCP), and the closest points between the robot $pos_{A}$ and the obstacles $pos_{B}$. Thus, the state space $S$ is defined as:
\begin{equation*}
	S=\left[ t, pos_{goal}, pos_{tcp}, vel_{tcp}, h, w, pos_{A}, pos_{B} \right] 
\end{equation*}

\subsubsection{Action}
Choosing the appropriate action greatly influences learning performance. Typically, there are two options for defining a robot's action space. One approach is to define the action space in the joint space of the robot, which requires calculating inverse kinematics (IK) at each time step since the positions of goals and obstacles are defined in Cartesian space. However, considering that many robots on the market today come with built-in inverse kinematics, training a policy based on Cartesian space makes it easier to transfer to other robots. Therefore, we defined the action space $A$ as:
\begin{equation*}
	A=\left[ \varDelta x, \varDelta y, \varDelta z, \varDelta g \right] 
\end{equation*}
where $\varDelta x$, $\varDelta y$ and $\varDelta z$ represent the relative motion along the $x$, $y$ and $z$ axes in Pybullet world coordinates, and $\varDelta g$ is the gripper command. All four components are bounded to $\left[-1,1\right]$. The gripper is not actuated by the reaching task, but it forms part of the action vector and therefore enters the effort term $r_a$ below.

\subsubsection{Reward Function}
To achieve the goal of reaching the target while avoiding obstacles, the agent has two separate objectives: (i) minimizing the distance between the end-effector and the goal position, and (ii) maximizing the distance between the end-effector and obstacles. Therefore, the reward function can be designed as follows:
\begin{equation*}
	r=c_1r_g+c_2r_a+c_3r_o
\end{equation*}
where $r_g$ represents the reward provided by the change in distance between the end-effector and the goal position:
\begin{equation*}
	r_{g_t}=\left\| pos_{goal_{t-1}} - pos_{tcp_{t-1}}  \right\| - \left\| pos_{goal_t} - pos_{tcp_t}  \right\| 
\end{equation*}
where $\left\| pos_{goal_t} - pos_{tcp_t}  \right\| $ is the Euclidean distance between the goal and the robot's end-effector at step $t$. $r_a$ is an effort term penalising the magnitude of the commanded action:
\begin{equation*}
	r_{a_t}=-\left\| a_t \right\| ^2 
\end{equation*}
$r_o$ represents the penalty when the robot gets too close to obstacles:
\begin{equation*}
	r_{o_t}=\begin{cases}
		-1 \ \ \    if\,\,d\left( pos_{A_t},pos_{B_t} \right) < d_{thre}\\
		0 \ \ \ \   \    if\,\,d\left( pos_{A_t},pos_{B_t} \right) \geqslant d_{thre}\\
	\end{cases}
\end{equation*}
where $d_{thre}=0.1$~m is the clearance threshold between the robot and the obstacles. The same penalty is applied independently to the fixed wall bounding the workspace and to the obstacle, so a state that violates both clearances is penalised twice. In addition, once contact is detected the environment latches an unsafe flag for the remainder of the episode, and a further penalty is subtracted from both proximity terms at every subsequent step rather than only at the step of contact. $c_1$, $c_2$ and $c_3$ are weights that balance the influence of the different reward components, listed in Table~\ref{table}, were selected by manual trial and error during framework development and held fixed for the reported experiments. With $c_1=500$ and $c_3=10$, each active obstacle-proximity term incurs a penalty equivalent in magnitude to the reward for a $0.02$~m reduction in goal distance. These settings therefore balance progress toward the goal against obstacle proximity, without imposing a hard safety constraint.
	
The action-magnitude weight was set to $c_2=0$, so the effort term is inactive and the reward
	used in Section~\ref{sec5} comprises only the goal-progress and obstacle terms. Command limits are
	enforced separately: the three translational action components are clipped to $[-1,1]$ and scaled
	to Cartesian target increments of at most approximately $0.042$~m per axis per step before
	workspace clipping. These limits bound the commanded increments, whereas a positive effort weight
	would encourage smaller actions within those limits and introduce an additional trade-off with goal
	progress.

\subsection{Integrating Human Demonstrations}
In this section, we provide details on how to collect human demonstrations using VR. We then propose our dual actor framework, which combines Imitation Learning and a Soft Actor Critic so that demonstrations are exploited without adding a direct imitation loss to the RL actor.

\subsubsection{Demonstration Collection} 
\begin{figure}[htbp]
	\centering
	\begin{subfigure}{0.99\linewidth}
		\centering
		\includegraphics[width=0.98\linewidth]{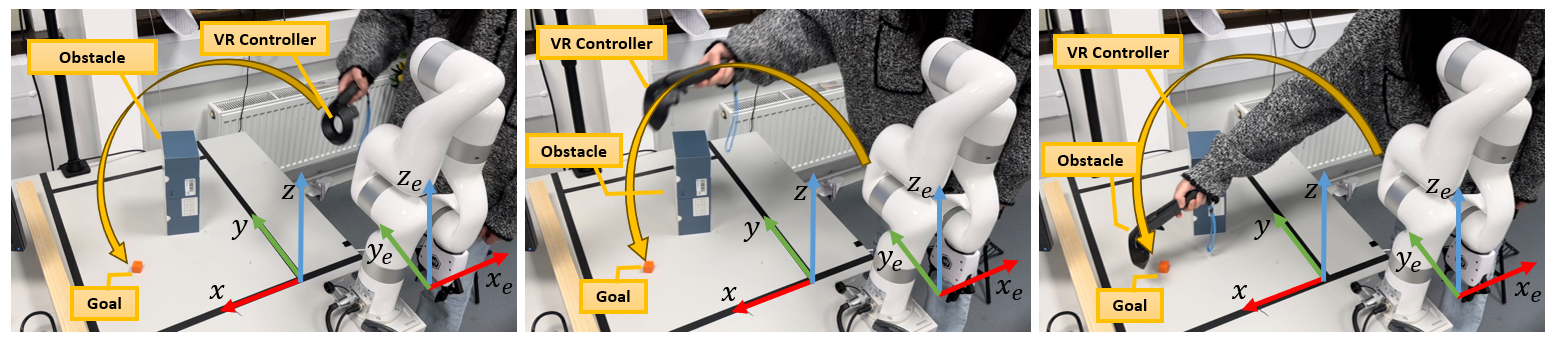}
		\caption{Physical world}
		\label{data_physical}
	\end{subfigure}
	\centering
	\begin{subfigure}{0.99\linewidth}
		\centering
		\includegraphics[width=0.98\linewidth]{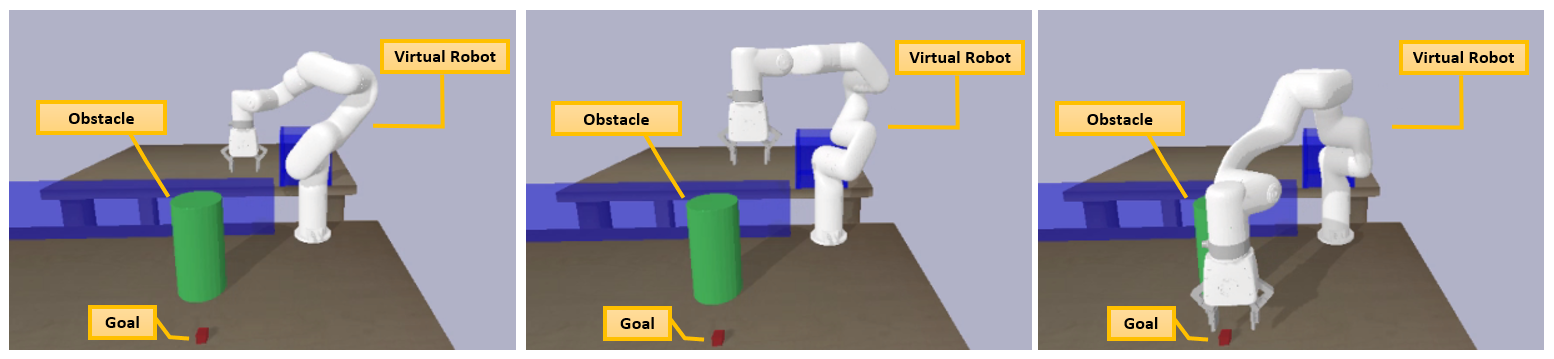}
		\caption{Digital world}
		\label{data_digital}
	\end{subfigure}
	\caption{Collecting human demonstrations by VR controllers.}
	\label{data}
\end{figure}

The HTC VIVE VR headset was used to collect human motion. As shown in Fig. \ref{data}, instead of controlling a real robot to complete the task and collect data, humans use a VR controller to guide the virtual robot around obstacles toward a goal. This method leverages the advantages of VR to collect data safely and intuitively. Additionally, since the DT is developed in conjunction with OpenAI Gym and Pybullet, each action can be labeled with a reward as transition $\left(s_t,a_t,s_{t+1},r_t\right)$. 

\subsubsection{Dual Actor Framework}
The main difference from existing ways of combining demonstrations with RL is \emph{where} the demonstrations enter the learner. Methods that add an IL term to the actor loss \cite{RN439,RN442,RN440,RN441} constrain the policy gradient, so non-optimal demonstrations may bias the actor when the critic misjudges their value. A Q-Filter \cite{RN439} disables imitation when the critic rates the actor's action higher, but this per-sample decision depends on the critic's accuracy. Our framework adds no direct imitation loss to the RL actor. Instead, a separate behaviour-cloning (BC) actor proposes an alternative action. At each update, the critic evaluates both actors and the bootstrapped target favours the path with the higher estimated value (Eq.~(\ref{da3})); the BC path retains a weight $v$ even when its estimated value is lower. Demonstrations therefore affect the RL actor only through the critic, allowing it to explore beyond the demonstrator.

We store agent interactions in replay buffer $\mathcal{R}$ and human demonstrations in a separate buffer $\mathcal{R}_D$. Like Twin Delayed Deep Deterministic Policy Gradient (TD3) and SAC, our framework uses two critics and the lower Q-value estimate to limit overestimation. The second actor learns only from $\mathcal{R}_D$ through behaviour cloning (BC), with objective $J_{BC}(\xi)$:

\begin{equation}
		J_{BC}\left( \xi \right) =\sum_{i=1}^{N_{\mathcal{R} _D}}{\left\| \pi_{BC}\left( s_i;\xi \right) -a_i \right\| ^2}
		\label{da1}
\end{equation}
where $a_i$ is the action from human demonstrations. In each mini-batch, we sample additional $N_{\mathcal{R} _D}$ transitions to update the second BC actor, and double actors generate double paths. According to Eq. (\ref{sac2}), assuming we have two critic networks  \(Q_{\theta_1}(s_t, a_t)\) and  \(Q_{\theta_2}(s_t, a_t)\), the value function is updated via:
\begin{equation}
	\hat{V}\left( s_{t+1} \right) \gets \min_{i=1,2} Q_{\theta _i}\left( s_{t+1},a_{t+1} \right) 
	\label{da2}
\end{equation}
Since we now have two actors, it is natural to favour the path with the higher value for updating. This approach helps the agent exploit the BC actor's path more while exploring further when the agent's performance surpasses that of the human. For brevity, let $Q_{\theta}^{\min}\left( s,\pi \right) =\min_{i=1,2} Q_{\theta _i}\left( s,\tilde{a} \right)-\alpha\log\pi\left(\tilde{a}|s\right)$, with $\tilde{a}\sim\pi\left(\cdot|s\right)$, denote the entropy-regularized value of following actor $\pi$ from state $s$, as in Eq.~(\ref{sac2}). The value function is then updated via:

\begin{equation}
	\begin{aligned}
	\hat{V}( s_{t+1} ) \gets {}&( 1-v ) \max \{ Q_{\theta}^{\min}( s_{t+1},\pi _{\phi}) , Q_{\theta}^{\min}( s_{t+1},\pi _{BC} ) \} \\
	&\quad+v\min \{ Q_{\theta}^{\min}( s_{t+1},\pi _{\phi} ) , Q_{\theta}^{\min}( s_{t+1},\pi _{BC} ) \}
	\end{aligned}
	\label{da3}
\end{equation}
where $v \in \mathbb{R}$ and it satisfies $0<v<1$. An alternative form replaces the second term of Eq.~(\ref{da3}) by $v\,Q_{\theta}^{\min}\left( s_{t+1},\pi _{BC} \right)$, so that only a positive estimated advantage of the RL actor over the BC actor is discounted. Both forms were evaluated with $v=0.2$ and did not differ detectably (Section~\ref{sec:quant}).


\section{Comparative Experiments and Discussion}\label{sec5}
\subsection{Task Description}
We use obstacle avoidance as a proof of concept for the online training framework. The Xarm5 must reach cube goals while avoiding box obstacles. If an attempt fails or is unsafe in the DT, the agent resumes training until it completes the task in simulation.

Experiments ran on Ubuntu 20.04 with an Intel Core i9-9960X CPU, 32 GB of memory, and an NVIDIA GeForce RTX 4070 Ti GPU. Before task execution, the agent was pre-trained with randomly positioned goals and a small obstacle. We loaded the Xarm5 URDF into PyBullet and used its inverse-kinematics solver to map agent actions to joint poses for the robot's five degrees of freedom. 

\begin{table}[h]
	\caption{Provenance of the demonstration sets.}
	\label{tab:demosource}
	\centering
	\footnotesize
	\setlength{\tabcolsep}{3.5pt}
	\begin{tabular}{lllrr}
		\toprule
		\makecell[l]{Demonstration\\set} & Source & Collection & \makecell{Traj.} & \makecell{Unique\\transitions} \\
		\midrule
		Optimal          & Expert policy   & \makecell[l]{Deterministic\\rollout}          & 10 & 330 \\
		Non-optimal      & Trained policy  & \makecell[l]{Stochastic\\rollout}             & 1  & 50  \\
		Human, complete  & Human operator  & \makecell[l]{VR controller,\\DT} & 10 & 433 \\
		Human, partial   & Human operator  & \makecell[l]{VR controller,\\DT} & 10 & 500 \\
		\bottomrule
	\end{tabular}
\end{table}

Table~\ref{tab:demosource} summarises the four demonstration datasets used in this study and their collection methods. The optimal and non-optimal sets are both generated by rolling out a trained policy, deterministically
and stochastically respectively; the non-optimal set therefore represents demonstrations of \emph{controlled}
sub-optimality rather than human error. The buffer stored for that condition repeats a single trajectory ten
times, so it contains 50 distinct transitions rather than the 500 stored entries. The two human sets were
recorded in the same scene by an operator guiding the virtual robot with the VR controller, and every
transition was labelled with a reward automatically by the DT. In the complete set the operator
guided the end-effector around the obstacle to the goal; in the partial set the operator guided it past the
obstacle but not onto the goal, so that none of these trajectories completes the task. Their collection
is described in Section~\ref{sec:effort}.

Note that, in addition to implementing our dual actor SAC in the experiment, we also implement the method proposed in \cite{RN439} as a comparative benchmark, which addresses a similar problem to this paper. However, their approach leverages the BC loss as an auxiliary loss to the original actor loss:
$$
L_{RL}=\lambda _1J_{\pi}\left( \phi \right) +\lambda _2 L_{BC}
$$
where $L_{BC}$ is the behaviour-cloning loss, filtered by a Q-Filter to handle suboptimal demonstrations:
\begin{equation}
\begin{aligned}
L_{BC}=\frac{1}{N_D}\sum_{i=1}^{N_D} & H\left(Q^{\min}(s_i,a_i)-Q^{\min}(s_i,\pi (s_i))\right)\\
 & \cdot \frac{1}{d_a}\left\| \pi (s_i|\theta _{\pi})-a_i \right\| ^2
\end{aligned}
\label{bc}
\end{equation}
where $Q^{\min}$ is the minimum of the two online critics (evaluated without gradient and without an entropy term, unlike the soft target values of Eq.~(\ref{da3})), $\pi(s_i)$ is the deterministic action of the actor (the squashed mean of its Gaussian), $d_a$ is the action dimension, and $H$ is a step function with $H(x)=1$ for $x>0$ and $H(x)=0$ otherwise, applied to each demonstration individually. Without the Q-Filter, $H\equiv1$.

\subsection{Agent Retraining}
After the agent learns to avoid a small obstacle, we replace it with a larger one. The pre-trained policy then collides while reaching for the goal; the supplementary video shows the DT and physical robot before and after retraining. PyBullet’s collision detector supplies only a safety flag, helping prevent unsafe simulated attempts from being executed on the robot. During training, the agent instead observes its shortest distance to the obstacle to maintain clearance.

\noindent\textbf{Video:} \href{https://youtu.be/pn2uJ113oG8}{\texttt{youtu.be/pn2uJ113oG8}}.

The ROS launches the retraining node as soon as it receives the flag message of task failure. The retraining experiments below resume from the saved pre-trained model, which has completed $1.029\times10^6$ interaction steps, and run for $1.5\times10^5$ additional steps, about 15\% of the pre-training interaction.

\begin{figure}[htbp]
	\centering
	\begin{subfigure}{0.8\linewidth}
		\centering
		\includegraphics[width=1\linewidth]{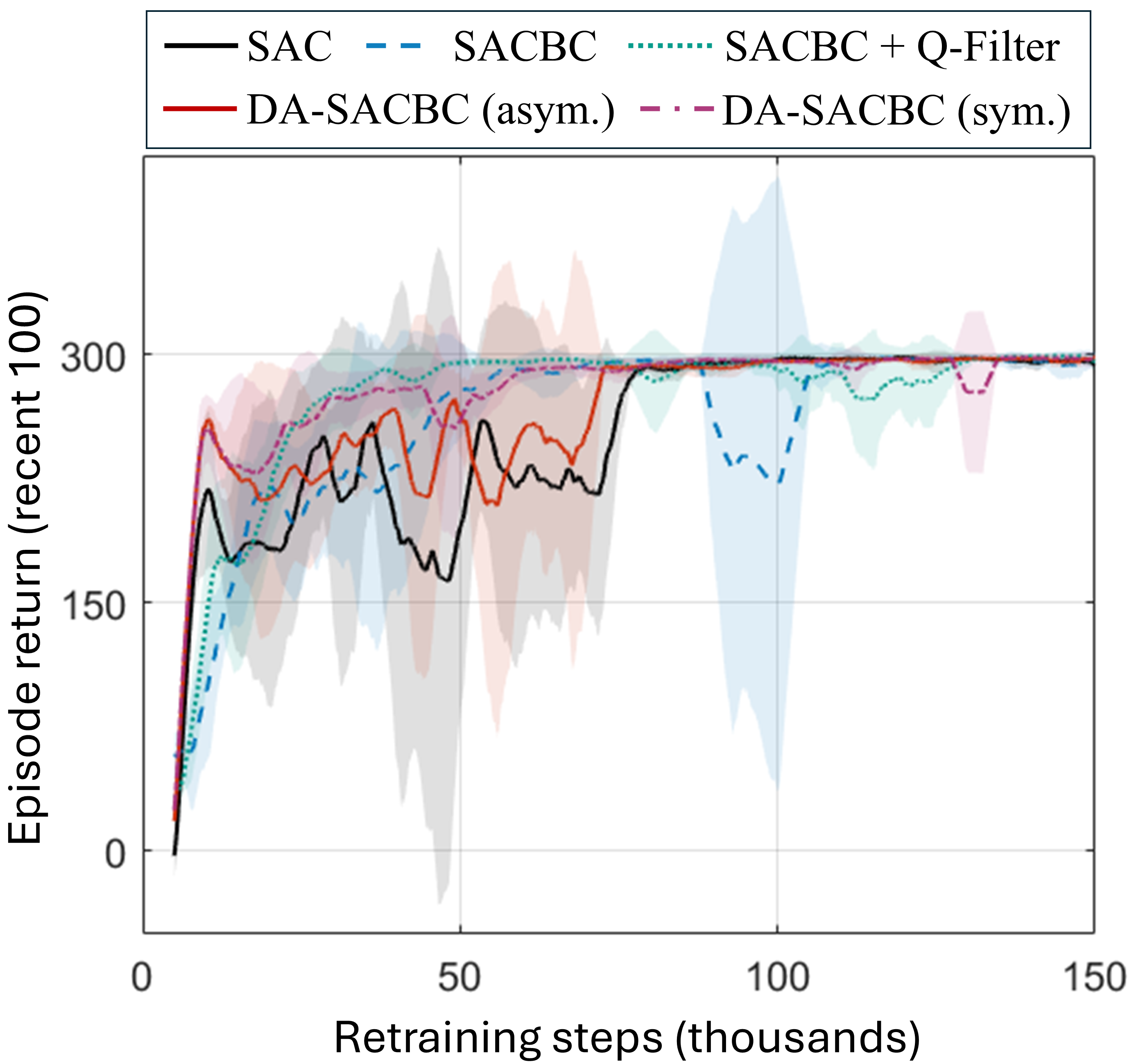}
		\caption{Return}
		\label{secondrun_optima_rew}
	\end{subfigure}
	\centering
	\begin{subfigure}{0.8\linewidth}
		\centering
		\includegraphics[width=1\linewidth]{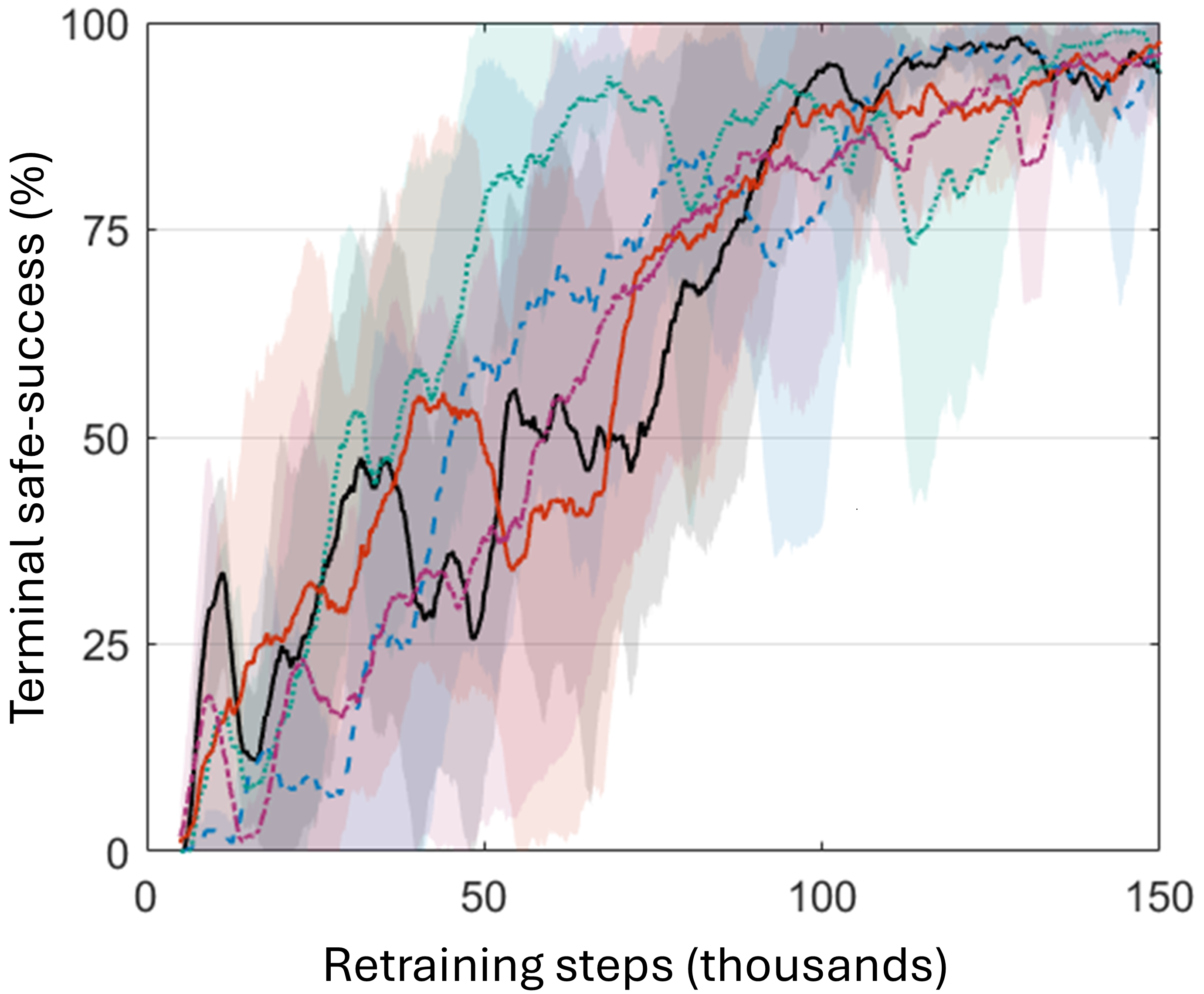}
		\caption{Success rate}
		\label{secondrun_optima_suc}
	\end{subfigure}
	\caption{Retraining with optimal demonstrations (policy-generated in Table~\ref{tab:demosource}). Lines: mean over six paired seeds; shaded: $\pm$ one standard deviation across seeds, truncated to [0,100]\% for success. (a) Return and (b) Terminal safe-success rate over the most recent 100 training episodes.}
	\label{optimal}
\end{figure}

Fig.~\ref{optimal} shows retraining in the nominal scene with the two policy-generated demonstration sets of Table~\ref{tab:demosource}. All methods resume from the same pre-trained checkpoint and replay buffer and train for $1.5\times10^5$ additional steps; each curve is the mean over six paired random seeds and the shaded region is $\pm$ one standard deviation across seeds. SACBC denotes SAC with the auxiliary BC loss of Eq.~(\ref{bc}) without filtering, SACBC + Q-Filter applies the filter \cite{RN439}, and DA-SACBC is the proposed dual actor framework, shown for both target forms discussed in Section~\ref{sec4}. With optimal demonstrations, the mean final training success of every method exceeds 90\%.

\begin{figure}[htbp]
	\centering
	\begin{subfigure}{0.8\linewidth}
		\centering
		\includegraphics[width=1\linewidth]{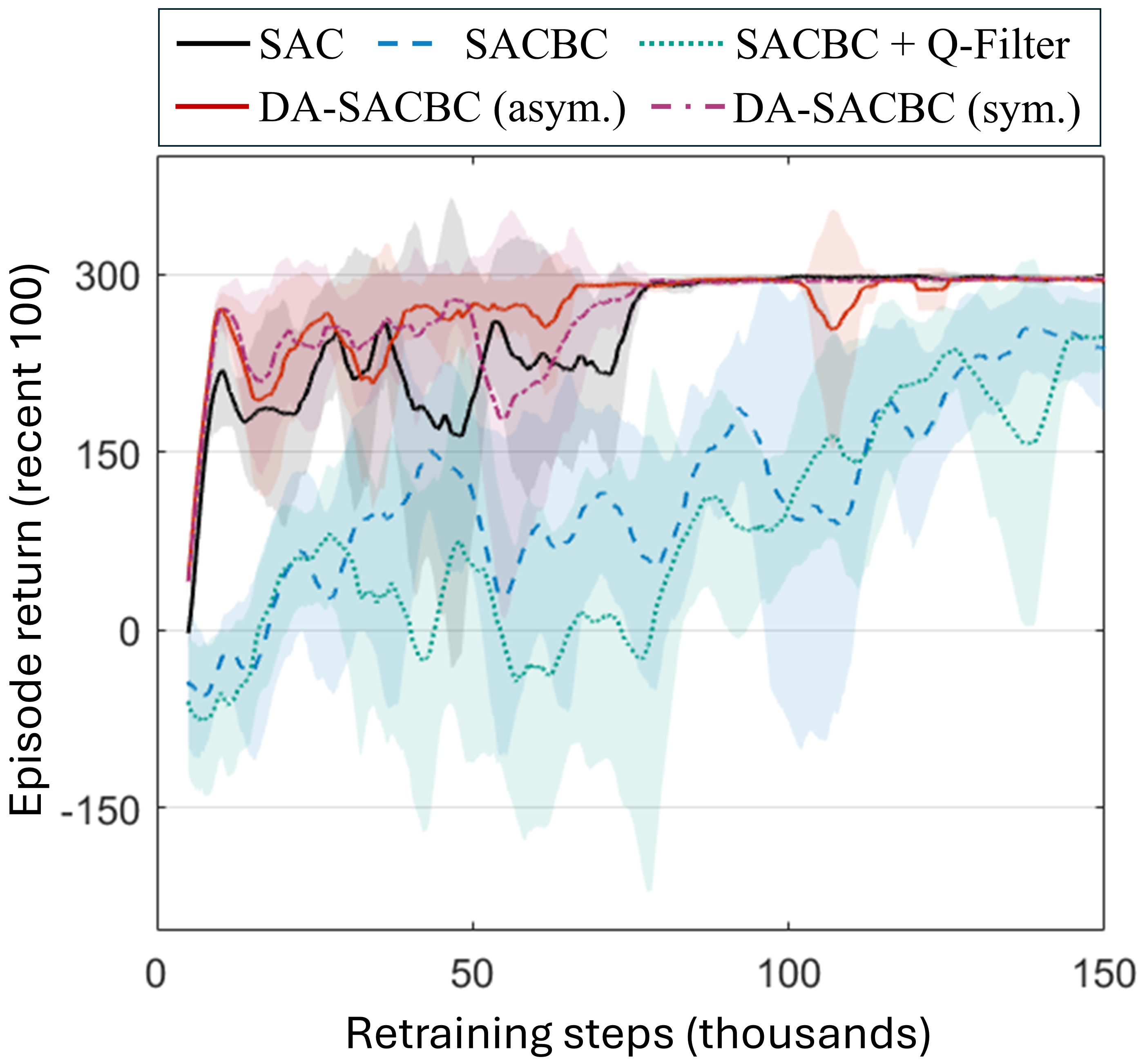}
		\caption{Return}
		\label{secondrun_nonoptima_rew}
	\end{subfigure}
	\centering
	\begin{subfigure}{0.8\linewidth}
		\centering
		\includegraphics[width=1\linewidth]{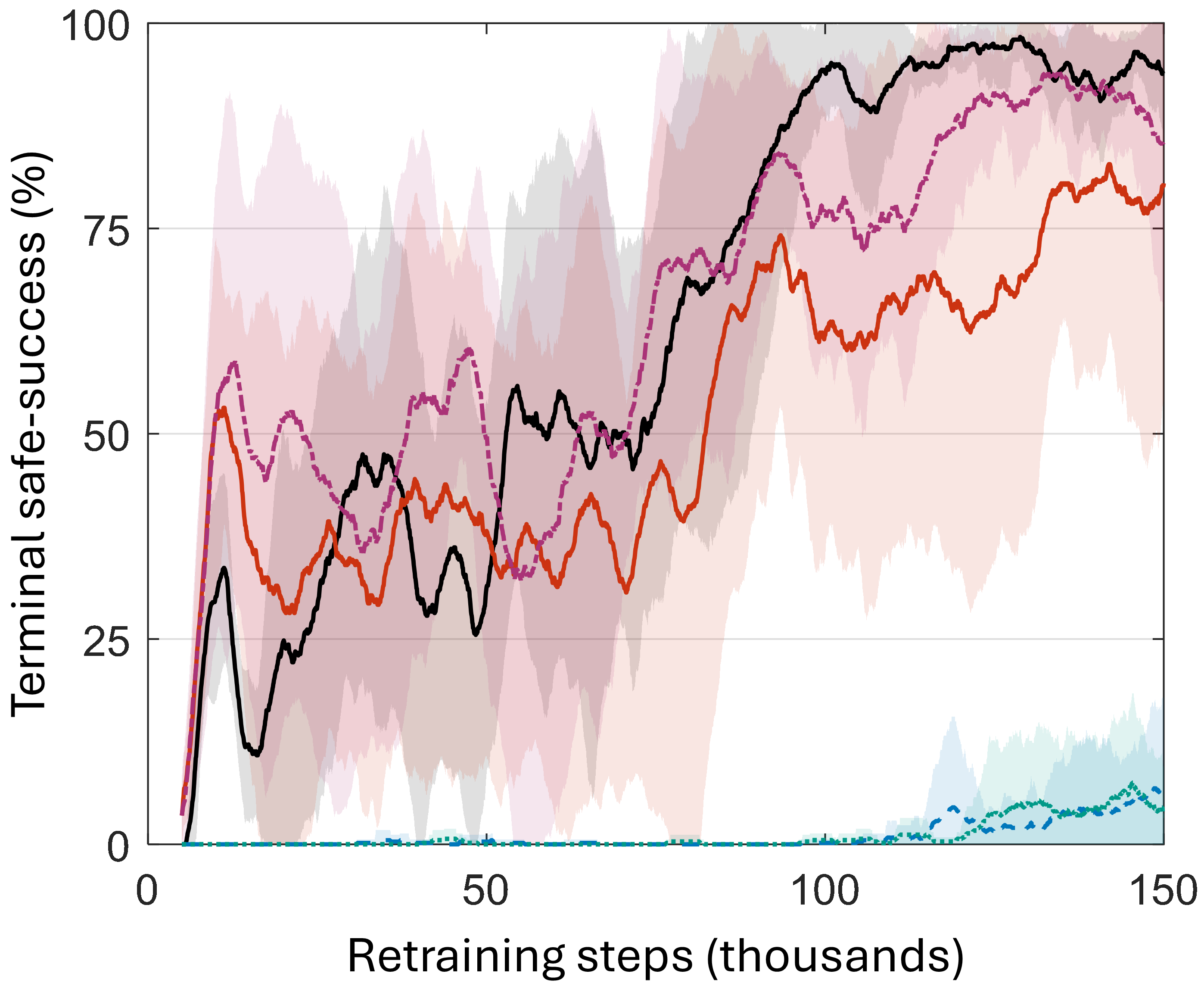}
		\caption{Success rate}
		\label{secondrun_nonoptima_suc}
	\end{subfigure}
	\caption{Retraining with non-optimal demonstrations (policy-generated in Table~\ref{tab:demosource}). Lines: mean over six paired seeds; shaded: $\pm$ one standard deviation across seeds, truncated to [0,100]\% for success. (a) Return and (b) Terminal safe-success rate over the most recent 100 training episodes.}
	\label{second}
\end{figure}

With non-optimal demonstrations, the results differ, as shown in Fig.~\ref{second}. For both methods that add an imitation term to the actor loss, the mean training success remains below 8\% throughout retraining, and the Q-Filter does not prevent this. This is consistent with the imitation term pulling the actor towards demonstrated behaviour that does not complete the task. The experiment compares complete algorithms, however, and does not isolate this mechanism: the baselines also mix demonstrations into the critic and actor mini-batches, whereas DA-SACBC trains its critic and RL actor on interaction data only and uses the demonstrations solely to train the BC actor. The mean success rate of DA-SACBC rises above 80\%, although only three of six seeds per target form meet the convergence criterion of Section~\ref{sec:quant}. DA-SACBC also has a head start over SAC without demonstrations on average: averaged over six seeds, its success rate over the first $5\times10^4$ retraining steps is 34.6\% and 44.1\% for the two target forms, against 28.3\% for SAC, and its mean return is 225.8 and 229.1, against 183.6. SAC's mean training success rate overtakes both DA-SACBC forms later in retraining. A reason for the faster start is that, through the weight $v$ in Eq.~(\ref{da3}), the BC actor's soft value contributes to the bootstrapped critic targets even when the RL path has the higher estimated value. As the BC actor learns from the demonstrations, this provides indirect guidance to the critic, whereas SAC learns from interaction replay, including the retained pre-training experience, without demonstrations.

After retraining, the robot autonomously avoids the larger obstacle, as shown in the supplementary video. Table~\ref{table} gives the hyperparameters used for training.

\begin{table}[htbp]
	\caption{Hyperparameters setup.}
	\centering
	\footnotesize
	\setlength{\tabcolsep}{4pt}
	\begin{tabular}{llr}
		\hline
		\hline
		\multicolumn{2}{l}{\textbf{Hyperparameter}}              &\textbf{Value}\\
		\hline
		\multicolumn{2}{l}{Shared}                      &  \\
		& Learning rate             & 0.0003               \\
		& Batch size                & 256                  \\
		& Optimizer                 & Adam                 \\
		& Actor network             & $\left(256,256\right)$\\
		& Critic network            & $\left(256,256\right)$\\
		& Activation function       & ReLU              \\
		& Buffer size               & 1000000              \\
		& Discount factor           & 0.99                 \\
		& \makecell[l]{Demonstration buffer\\(stored transitions)} & \makecell[l]{policy 330 / 500;\\human 433 / 500} \\
		& Reward weights $\left(c_1,c_2,c_3\right)$ & $\left(500,\ 0,\ 10\right)$ \\
		& Obstacle clearance $d_{thre}$ & 0.1 m \\
		\hline
		\multicolumn{2}{l}{Loss SACBC} & \multicolumn{1}{l}{} \\
		& $\lambda_1$                   & 1                    \\
		& $\lambda_2$ & 0.87 \\
		& Demonstration fraction per mini-batch & 0.5 \\
		\hline
		\multicolumn{2}{l}{Dual Actor SACBC} & \multicolumn{1}{l}{} \\
		& $v$ & 0.2\\
		& BC-actor demonstration mini-batch & 256 \\
		& BC actor network                     & $\left(256,256\right)$ \\
		\hline
		\hline
	\end{tabular}
	\label{table}
\end{table}

\subsection{Quantitative Comparison}\label{sec:quant}
Table~\ref{tab:metrics} reports scalar metrics for the runs in Fig.~\ref{optimal} and Fig.~\ref{second}: six paired seeds per method, mean $\pm$ standard deviation across seeds. A run is counted as converged at the first retraining step at which the success rate over the most recent 100 training episodes reaches 90\% and stays at or above 90\% until the end of the $1.5\times10^5$-step budget; the convergence column gives the number of converged seeds and the mean over those seeds. Training success is the terminal safe-success rate, and average return the mean episode return, over the final 100 training episodes of each run. Evaluation success is measured separately, over 100 deterministic evaluation episodes of each final policy. Contact episodes count the training episodes with at least one sampled contact between a gripper finger and the wall or the obstacle. All methods use the same $1.5\times10^5$ interaction steps; with four runs sharing one GPU, each of the 24 runs timed under these conditions took 1.85--2.07~h.

With optimal demonstrations, every method converges in at least five of six seeds and reaches a comparable final success rate; paired comparisons of the convergence step against SAC show no consistent advantage, so none of the demonstration-based methods is shown to accelerate retraining relative to SAC. With non-optimal demonstrations, SACBC and SACBC + Q-Filter converge in none of the six seeds and end with training success rates of 7.3\% and 3.5\%, whereas DA-SACBC reaches 80.5\% and 85.0\% for the two target forms and achieves a higher final training success than both baselines in every paired seed. In the independent evaluation of the final policies, both DA-SACBC forms again retain a higher mean success than both imitation-loss baselines: with non-optimal demonstrations, the DA-SACBC policies reach a mean deterministic evaluation success of 83.3\% and 100.0\%, against 0.0\% for SACBC and 16.7\% for SACBC + Q-Filter. DA-SACBC also records the lowest mean number of sampled contact episodes under both demonstration sets; with non-optimal demonstrations these are 149 and 171 of 3000 training episodes, against 728 and 831 for the two imitation-loss baselines and 422 for SAC. These counts are sampled finger contacts during training rather than collisions of the final policy, and they may also depend on the trajectories visited and the time spent near the obstacle. The two target forms of Eq.~(\ref{da3}) did not differ detectably in convergence with non-optimal demonstrations.

\begin{table*}[t]
	\caption{Retraining metrics in the nominal scene (six paired seeds; mean $\pm$ standard deviation across seeds). The SAC baseline uses no demonstrations and is shared by both demonstration sets. DA-SACBC (asym.) and (sym.): asymmetric and symmetric target forms of Eq.~(\ref{da3}).}
	\label{tab:metrics}
	\centering
	\footnotesize\ifrev\color{black}\fi
	\begin{tabular}{llccccc}
		\toprule
		Demos & Method & \makecell{Converged seeds;\\steps ($10^3$)} & \makecell{Training\\success (\%)} & \makecell{Evaluation\\success (\%)} & \makecell{Average\\return} & \makecell{Sampled contact\\episodes (of 3000)} \\
		\midrule
		None & SAC & 5/6; $118.7\pm27.8$ & $93.8\pm7.2$ & $100.0\pm0.0$ & $293.5\pm5.9$ & $422\pm159$ \\
		\midrule
		\multirow{4}{*}{Optimal} & SACBC & 5/6; $95.1\pm39.5$ & $97.5\pm4.7$ & $100.0\pm0.0$ & $297.4\pm4.4$ & $671\pm166$ \\
		 & SACBC + Q-Filter & 5/6; $114.9\pm27.1$ & $94.0\pm8.0$ & $83.3\pm40.8$ & $295.6\pm5.2$ & $539\pm126$ \\
		 & DA-SACBC (asym.) & 6/6; $123.3\pm23.3$ & $97.7\pm2.5$ & $100.0\pm0.0$ & $297.9\pm1.1$ & $249\pm154$ \\
		 & DA-SACBC (sym.) & 5/6; $121.6\pm21.6$ & $96.3\pm6.1$ & $100.0\pm0.0$ & $296.9\pm4.4$ & $220\pm115$ \\
		\midrule
		\multirow{4}{*}{Non-optimal} & SACBC & 0/6; -- & $7.3\pm10.5$ & $0.0\pm0.0$ & $237.2\pm53.2$ & $728\pm255$ \\
		 & SACBC + Q-Filter & 0/6; -- & $3.5\pm6.0$ & $16.7\pm40.8$ & $248.4\pm27.9$ & $831\pm724$ \\
		 & DA-SACBC (asym.) & 3/6; $117.5\pm37.2$ & $80.5\pm26.5$ & $83.3\pm40.8$ & $294.8\pm3.8$ & $149\pm23$ \\
		 & DA-SACBC (sym.) & 3/6; $123.9\pm16.9$ & $85.0\pm19.6$ & $100.0\pm0.0$ & $295.8\pm1.7$ & $171\pm71$ \\
		\bottomrule
	\end{tabular}
\end{table*}

\subsection{Retraining with Human VR Demonstrations}\label{sec:human}
The comparison of Section~\ref{sec:quant} was repeated with the two human demonstration sets of
Table~\ref{tab:demosource}, keeping every other setting unchanged: the same pre-trained checkpoint and
retained replay buffer, the same $1.5\times10^5$ retraining steps, hyperparameters (Table~\ref{table}),
six seeds and evaluation protocol. SAC without demonstrations is the same set of runs as in
Table~\ref{tab:metrics}.

\begin{table*}[t]
	\caption{Retraining with human VR demonstrations (mean $\pm$ SD over six seeds). Columns as in Table~\ref{tab:metrics}.} 
	\label{tab:human}
	\centering
	\footnotesize\ifrev\color{black}\fi
	\begin{tabular}{llccccc}
		\toprule
		Demonstrations & Method & \makecell{Converged seeds;\\steps ($10^3$)} & \makecell{Training\\success (\%)} & \makecell{Evaluation\\success (\%)} & \makecell{Average\\return} & \makecell{Sampled contact\\episodes (of 3000)} \\
		\midrule
		None & SAC & 5/6; $118.7\pm27.8$ & $93.8\pm7.2$ & $100.0\pm0.0$ & $293.5\pm5.9$ & $422\pm159$ \\
		\midrule
		\multirow{4}{*}{Human, complete} & SACBC & 0/6; -- & $31.7\pm38.0$ & $33.3\pm51.6$ & $285.6\pm12.4$ & $514\pm178$ \\
		 & SACBC + Q-Filter & 1/6; $149.35$ & $36.8\pm36.2$ & $50.0\pm54.8$ & $291.1\pm3.9$ & $425\pm108$ \\
		 & DA-SACBC (asym.) & 6/6; $94.6\pm26.3$ & $99.5\pm0.8$ & $100.0\pm0.0$ & $299.2\pm0.8$ & $124\pm32$ \\
		 & DA-SACBC (sym.) & 6/6; $89.2\pm30.1$ & $99.0\pm1.7$ & $100.0\pm0.0$ & $299.2\pm0.6$ & $134\pm36$ \\
		\midrule
		\multirow{4}{*}{Human, partial} & SACBC & 0/6; -- & $8.0\pm18.6$ & $16.7\pm40.8$ & $148.9\pm123.0$ & $1303\pm398$ \\
		 & SACBC + Q-Filter & 0/6; -- & $0.3\pm0.8$ & $0.0\pm0.0$ & $-69.8\pm163.0$ & $1375\pm444$ \\
		 & DA-SACBC (asym.) & 5/6; $105.5\pm29.8$ & $82.3\pm37.0$ & $100.0\pm0.0$ & $287.4\pm25.9$ & $246\pm136$ \\
		 & DA-SACBC (sym.) & 3/6; $95.2\pm15.8$ & $76.2\pm24.7$ & $83.3\pm40.8$ & $291.6\pm7.0$ & $319\pm216$ \\
		\bottomrule
	\end{tabular}
\end{table*}

As shown in Table~\ref{tab:human}, with the complete human demonstrations, both forms of DA-SACBC converged in all six seeds and reached a
deterministic evaluation success of 100\%, whereas the two methods that add an imitation loss to the actor
converged in none and one of six seeds and reached 33.3\% and 50.0\%. With the partial demonstrations, which
never reach the goal, DA-SACBC reached 100\% and 83.3\% evaluation success, against 16.7\% for SACBC and 0\%
for SACBC + Q-Filter, and in each of the six paired seeds both DA-SACBC forms ended
retraining with a higher training success than either imitation-loss baseline. Under each demonstration set, DA-SACBC also recorded lower mean numbers of
sampled contact episodes during retraining than both imitation-loss baselines (124--134 against 425--514 with
the complete set, 246--319 against 1303--1375 with the partial set), and lower than SAC without
demonstrations (422). With the partial demonstrations, retraining of the two imitation-loss baselines was
also unstable: in 10 of their 12 runs the automatically tuned entropy coefficient rose from about 0.04 at the first logged point
of retraining to between 43 and 444 and the logged critic loss rose above $4\times10^{7}$, whereas in all 24 DA-SACBC runs the coefficient
stayed below 0.14.

\subsection{Generalization to More Complex Scenarios}\label{sec:gen}
In this section, we retrained in three further scenes in the DT, each derived from the nominal scene by one fixed geometric
change in simulation: \emph{S1}, a second obstacle, a 0.42~m pillar of radius 4.5~cm placed at $(-0.375, 0.32)$~m on the path
taken by the policies retrained in the nominal scene; \emph{S2}, the cylinder replaced by a tall box of
$7\times36\times32$~cm (width, depth, height) at the same horizontal location; and \emph{S3}, the goal moved from
$(-0.65, 0.23)$~m to $(-0.72, 0.05)$~m with the obstacle unchanged. In all three scenes the existing policies fail: the pre-trained policy and each of the six SAC policies retrained in
the nominal scene (Table~\ref{tab:metrics}) succeed in 0 of 100 deterministic evaluation episodes, whereas a scripted
waypoint route reaches the goal without a sampled contact between a finger and the wall or obstacle, motivating retraining of the tested policies.
The same retraining protocol as in Section~\ref{sec:quant} was then applied with SAC, SACBC and the asymmetric
DA-SACBC, six seeds each, from the same pre-trained checkpoint and retained replay buffer. Because the SAC teacher
did not meet the registered acceptance criterion in any of these scenes, the ten complete demonstrations per scene
came from a scripted waypoint route executed in the twin. In S2 and S3 we also retrained with \emph{partial} scripted demonstrations,
using the same learning and evaluation protocol: the controller stopped 0.29~m (S2) and 0.14~m (S3) short of the
goal and held position to the 50-step horizon. Table~\ref{tab:gen} reports the results, using the same metrics and definitions as
Table~\ref{tab:metrics}.

\begin{table*}[t]
	\caption{Retraining in three additional fixed scenes (six paired seeds; mean $\pm$ SD). Columns as in
	Table~\ref{tab:metrics}; DA-SACBC uses the asymmetric target.}
	\label{tab:gen}
	\centering
	\footnotesize\ifrev\color{black}\fi
	\begin{tabular}{lllccccc}
		\toprule
		Scene & Demos & Method & \makecell{Converged seeds;\\steps ($10^3$)} & \makecell{Training\\success (\%)} & \makecell{Evaluation\\success (\%)} & \makecell{Average\\return} & \makecell{Sampled contact\\episodes (of 3000)} \\
		\midrule
		\multirow{3}{*}{S1 pillar} & None & SAC & 0/6; -- & $15.0\pm12.1$ & $0.0\pm0.0$ & $274.7\pm11.4$ & $835\pm501$ \\
		 & Complete & SACBC & 0/6; -- & $19.5\pm13.9$ & $50.0\pm54.8$ & $273.1\pm10.9$ & $1004\pm430$ \\
		 & Complete & DA-SACBC & 2/6; $133.4\pm20.5$ & $62.0\pm34.6$ & $83.3\pm40.8$ & $284.8\pm11.3$ & $644\pm438$ \\
		\midrule
		\multirow{5}{*}{S2 box} & None & SAC & 1/6; $140.9$ & $24.7\pm34.7$ & $16.7\pm40.8$ & $254.6\pm28.9$ & $834\pm261$ \\
		 & Complete & SACBC & 4/6; $115.2\pm11.0$ & $94.2\pm7.7$ & $100.0\pm0.0$ & $297.8\pm1.3$ & $134\pm65$ \\
		 & Complete & DA-SACBC & 1/6; $137.8$ & $31.7\pm37.9$ & $16.7\pm40.8$ & $258.7\pm28.5$ & $785\pm332$ \\
		 & Partial & SACBC & 6/6; $121.8\pm22.3$ & $98.0\pm3.5$ & $100.0\pm0.0$ & $298.1\pm1.0$ & $133\pm46$ \\
		 & Partial & DA-SACBC & 0/6; -- & $22.5\pm20.2$ & $33.3\pm51.6$ & $233.2\pm31.3$ & $990\pm400$ \\
		\midrule
		\multirow{5}{*}{S3 moved goal} & None & SAC & 1/6; $142.9$ & $56.0\pm32.8$ & $66.7\pm51.6$ & $288.7\pm42.6$ & $420\pm223$ \\
		 & Complete & SACBC & 0/6; -- & $25.8\pm12.9$ & $33.3\pm51.6$ & $292.9\pm7.1$ & $488\pm278$ \\
		 & Complete & DA-SACBC & 3/6; $141.2\pm6.7$ & $64.3\pm42.5$ & $66.7\pm51.6$ & $304.1\pm10.2$ & $222\pm55$ \\
		 & Partial & SACBC & 0/6; -- & $32.0\pm37.3$ & $33.3\pm51.6$ & $252.3\pm98.7$ & $866\pm604$ \\
		 & Partial & DA-SACBC & 2/6; $149.6\pm0.2$ & $73.5\pm20.9$ & $66.7\pm51.6$ & $305.4\pm5.6$ & $530\pm117$ \\
		\bottomrule
	\end{tabular}
\end{table*}

The outcome depends on the scene. With the pillar (S1), DA-SACBC was the only method to converge in any seed and
reached a mean deterministic evaluation success of 83\%, against 50\% for SACBC and 0\% for SAC; its final training
success exceeded SAC's in all six paired seeds and SACBC's in five. With the moved goal (S3) and complete
demonstrations, DA-SACBC and SAC without demonstrations reached the same mean deterministic evaluation success
(67\%), both above SACBC (33\%), and DA-SACBC had the lowest mean number of sampled-contact episodes of the three
(222 against 420 and 488). With the tall box (S2) the order reversed: SACBC converged in four seeds and reached 100\%
deterministic evaluation success with 134 contact episodes on average, whereas DA-SACBC (17\%, 785 contact episodes)
had the same mean deterministic evaluation success as SAC without demonstrations; SACBC's final training success
exceeded DA-SACBC's in five of six seeds, with one tie. Replacing the complete demonstrations by partial ones
retained the DA--SACBC ordering in both scenes: SACBC still reached 100\% in S2, and DA-SACBC still exceeded SACBC
in mean deterministic evaluation success in S3 (67\% against 33\%); on final training success DA-SACBC was ahead of
SACBC in all six paired seeds in S3 and behind in all six in S2. Contact counts did change with partial
demonstrations.

\subsection{Human Effort and Demonstration Quality}\label{sec:effort}
The operator provides demonstrations only when the DT detects failure and need not program, reset, or supervise the robot during training. To measure this effort, one operator with occasional prior VR use collected both sets in Table~\ref{tab:demosource} during a 60~min session. Each attempt was limited to 50 steps (5~s at 10~Hz) and accepted only if it met the set's criterion without sampled wall or obstacle contact. For the complete set, 10 of 230 attempts were accepted (4.3\%) over 35~min; accepted trajectories lasted 39--50 steps. For the partial set, which omitted the precise final approach, 10 of 77 were accepted (13\%) over 10~min; accepted trajectories lasted 50 steps and ended 5.2--10.5~cm from the goal. Mean raw NASA-TLX workload across six subscales was 60.8 for the complete set and 46.7 for the partial set. For complete demonstrations, effort was highest (90), followed by mental demand and performance (75 each). The partial set therefore took fewer attempts and less time, while unsuccessful trials in the DT posed no collision risk to the physical robot. Section~\ref{sec:human} shows 83--100\% mean deterministic evaluation success with either set, including partial demonstrations that never reach the goal. These measurements describe one operator's effort in one session, not general usability, which requires a study with multiple operators.

\section{Conclusions}\label{sec6}
We presented a hardware-in-the-loop DT framework that uses camera feedback to keep the virtual workspace synchronized and previews obstacle-avoidance actions before physical execution. When the workspace changes, RL training resumes in the DT. A separate behaviour-cloning actor guides learning through the critic without adding a direct imitation loss to the RL actor, allowing it to explore beyond the demonstrator. With fixed non-optimal demonstrations, mean final training terminal safe-success exceeded 80\%, compared with below 8\% for two imitation-loss baselines. The final policies also had higher mean evaluation success and fewer sampled-contact episodes during retraining. With partial human VR demonstrations, mean deterministic evaluation success was 83--100\%, versus 0--17\% for the baselines. The framework currently assumes that relevant workspace changes appear in the RL observations, such as obstacle size; changes outside that representation and gaps in camera coverage may limit adaptation. Operator effort and workload were measured for one person in one session, so usability across users requires a dedicated study.

Future work will address obstacles that move during task execution. Their motion requires faster or predictive DT synchronization than the current perception pipeline and one-episode look-ahead, which assume a quasi-static scene. The state must include velocity estimates or short-horizon motion predictions, and the policy must be trained across a range of motions rather than retrained after every change. Because a trajectory judged safe in the DT may be unsafe by the time the physical robot executes it, contact-penalty rewards should be complemented by safe-RL mechanisms such as control barrier functions. Lidar could improve scene sensing, while force/torque sensing could help detect contact with co-workers. Moving obstacles also make demonstrations harder to time; we will integrate VR headsets into the DT so operators can see the virtual robot and obstacles from their own viewpoint.

\bibliographystyle{IEEEtran}
\bibliography{References_iros}

\end{document}